\documentclass[10pt,twocolumn,letterpaper]{article}

\usepackage{cvpr}              
\usepackage{pgfplots}
\usepgfplotslibrary{groupplots}
\pgfplotsset{compat=1.18}
\usepackage{algorithm}
\usepackage{adjustbox}
\usepackage{algorithmic}
\usepackage{pgfplotstable}
\usepackage{caption}
\usepackage{subcaption}
\usepackage{placeins} 

\definecolor{cvprblue}{rgb}{0.21,0.49,0.74}
\usepackage[pagebackref,breaklinks,colorlinks,allcolors=cvprblue]{hyperref}

\def\paperID{14414} 
\def\confName{CVPR}
\def\confYear{2025}

\title{RDDPM: Robust Denoising Diffusion Probabilistic Model for Unsupervised Anomaly Segmentation}

\author{Mehrdad Moradi, Kamran Paynabar\\
H. Milton Stewart School of Industrial and Systems Engineering, Georgia Institute of Technology\\
Atlanta, Georgia\\
{\tt\small mmoradi6@gatech.edu,kamran.paynabar@isye.gatech.edu}
\thanks{This work has been accepted to the ICCV 2025 Workshop on Vision-based Industrial InspectiON (VISION).}
}

\begin{document}
\maketitle
\begin{abstract}
Recent advancements in diffusion models have demonstrated significant success in unsupervised anomaly segmentation. For anomaly segmentation, these models are first trained on normal data; then, an anomalous image is noised to an intermediate step, and the normal image is reconstructed through backward diffusion. Unlike traditional statistical methods, diffusion models do not rely on specific assumptions about the data or target anomalies, making them versatile for use across different domains. However, diffusion models typically assume access to normal data for training, limiting their applicability in realistic settings.  In this paper, we propose novel robust denoising diffusion models for scenarios where only contaminated (i.e., a mix of normal and anomalous) unlabeled data is available. By casting maximum likelihood estimation of the data as a nonlinear regression problem, we reinterpret the denoising diffusion probabilistic model through a regression lens. Using robust regression, we derive a robust version of denoising diffusion probabilistic models. Our novel framework offers flexibility in constructing various robust diffusion models. Our experiments show that our approach outperforms current state of the art diffusion models, for unsupervised anomaly segmentation when only contaminated data is available. Our method outperforms existing diffusion-based approaches, achieving up to 8.08\% higher AUROC and 10.37\% higher AUPRC on MVTec datasets.
The implementation code is available at: \href{https://github.com/mehrdadmoradi124/RDDPM}{https://github.com/mehrdadmoradi124/RDDPM}
\end{abstract}    
\section{Introduction}
\label{sec:intro}
\begin{figure}
     \centering
    \begin{subfigure}{0.3\linewidth}
         \includegraphics[width=0.9\linewidth]{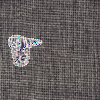}
         \caption{}
         \label{fig:sparse}
     \end{subfigure}
    \begin{subfigure}{0.3\linewidth}
         \includegraphics[width=0.9\linewidth]{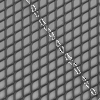}
         \caption{}
         \label{fig:sparse}
     \end{subfigure}
    \begin{subfigure}{0.3\linewidth}
         \includegraphics[width=0.9\linewidth]{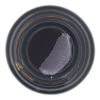}
         \caption{}
         \label{fig:sparse}
     \end{subfigure}
     \begin{subfigure}{0.3\linewidth}
         \includegraphics[width=0.9\linewidth]{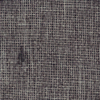}
         \caption{}
         \label{fig:anomalous}
     \end{subfigure}
     \begin{subfigure}{0.3\linewidth}
         \includegraphics[width=0.9\linewidth]{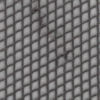}
         \caption{}
         \label{fig:smooth}
     \end{subfigure}
     \begin{subfigure}{0.3\linewidth}
         \includegraphics[width=0.9\linewidth]{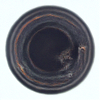}
         \caption{}
         \label{fig:sparse}
     \end{subfigure}
          \begin{subfigure}{0.3\linewidth}
         \includegraphics[width=0.9\linewidth]{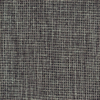}
         \caption{}
         \label{fig:anomalous}
     \end{subfigure}
     \begin{subfigure}{0.3\linewidth}
         \includegraphics[width=0.9\linewidth]{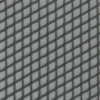}
         \caption{}
         \label{fig:smooth}
     \end{subfigure}
     \begin{subfigure}{0.3\linewidth}
         \includegraphics[width=0.9\linewidth]{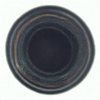}
         \caption{}
         \label{fig:sparse}
     \end{subfigure}
        \caption{a-c: Anomalous samples of carpet (a), grid (b), and bottle (c) with anomalies highlighted. d-f: Reconstructed images from DDPM trained on 20 percent contaminated data. g-i: Anomaly free reconstructed images with RDDPM in 20 percent contaminated data. }
        \label{fig:DDPM contaminated}
\end{figure}
Diffusion models have demonstrated tremendous success in image synthesis and density estimation \cite{rombach2022high_diff6, dhariwal2021diffusion_diff5,kingma2023variational_diff12}. Consequently, reconstruction-based anomaly detection and segmentation using diffusion models have gained significant success \cite{DIFFUSION_AD_dynamic_noise, DTE_DIFFUSION_AD, Real_net_DIFFUSION_AD, wyatt_anoddpm_2022_DIFFUSION_AD, mousakhan_anomaly_2023_DIFFUSION_AD, zhang_diffusionad_2023_DIFFUSION_AD, iqbal_unsupervised_2023_DIFFUSION_AD, zhang_unsupervised_2023_diffusion_ad, fucka_transfusion_2024_diffusion_ad, imdiffusion_ad_Diffusion}. To apply diffusion models in anomaly segmentation, a model is trained on normal data. Anomalous data is then reconstructed to closely resemble normal data, resulting in an anomaly-free image reconstruction.
\\
However, the assumption of having access to normal data for training is not realistic in many manufacturing and biomedical contexts. Although DDPM-based methods for reconstruction of anomaly free images  are very powerful in learning complicated patterns in the data, as illustrated in \cref{fig:DDPM contaminated}, they fail when training data is contaminated. This results in a higher false alarm rate. This experiment was conducted on MVTec data set \cite{bergmann_mvtec_2019}, a widely used benchmark for unsupervised anomaly detection \cite{akcay2022anomalib_MVTec_support}, with DDPM model.

To handle contaminated data, numerous matrix decomposition approaches have been proposed \cite{RPCA, SSD}. Robust Principal Component Analysis (RPCA) \cite{RPCA}, decomposes the image into low-rank and sparse components representing normal and anomaly part, respectively. These methods impose structural constraints on anomaly or normal background and employ optimization techniques to separate the components. However, these structural constraints can limit their effectiveness when applied to complex datasets. To mitigate this limitation, \cite{Robust_AE_AD} proposed to use an autoendoer as the low-rank normal component whithin RPCA to handle non-linear background. Autoencoders, however, have been shown to suffer from reconstruction quality issues \cite{zhang_diffusionad_2023_DIFFUSION_AD, wyatt_anoddpm_2022_DIFFUSION_AD}. Given the success of diffusion models in image synthesis \cite{dhariwal2021diffusion_diff5, rombach2022high_diff6}, there is a growing need to develop a diffusion model that is robust to outliers in the data for effective anomaly segmentation.\\
\cite{anomaly_rejection} introduced a rejection scheme in DDPM training algorithm, discarding data points with high residuals. \cite{sudo_huber} trained a  denoising score matching diffusion model with pseudo-Huber loss to reduce the impact of outliers on generated images. These approaches show initial promise for the development of robust diffusion models; however, they lack theoretical justification and concrete evidence to support their effectiveness.
 \\
 In this paper, we propose a novel framework for training a DDPM that is robust to outliers. We cast the problem as a nonlinear regression and replace the loss function with a statistically robust counterpart. This enables the model to learn the underlying data distribution without learning outliers. By introducing a robustness hyper-parameter, our model allows for adjusting robustness according to the problem setting and relevant domain knowledge. Our contributions are summarized below:
 \begin{itemize}
     \item We introduce a statistically equivalent formulation for DDPM, allowing us to reinterpret the model as a nonlinear regression problem. 
     \item We use robust functions to develop generalized versions of DDPM robust to training data contamination and outliers. 
     \item We introduce robustness parameter which can control the level of robustness, tunable for different settings.
\end{itemize}

\section{Related Work}
\label{sec:related_work}
\subsection{Diffusion Models}
Diffusion probabilistic models \cite{DDPM_2020} have shown strong performance in image synthesis \cite{rombach2022high_diff6, dhariwal2021diffusion_diff5} and density estimation \cite{kingma2023variational_diff12}. Training these models typically involves using a UNet architecture \cite{ronneberger2015unet_diff11} to predict the Gaussian noise added to the sampled image or gradient of the data distribution \cite{dhariwal2021diffusion_diff5, DDPM_2020, song2021score_diff10}.
\\One major issue with diffusion models is their high computational cost during inference. A body of the literature has worked on sampling efficiency \cite{kong2021fast_diff8, san2021noise_diff7, DDIM}. Additionally, there is some research on hierarchical approaches \cite{ho2021cascaded_diff_1}, and on generative modeling in the latent space \cite{rombach2022high_diff6, vahdat2021score_diff9} to address training and evaluation cost. \\
Our method, RDDPM, is a generalization of DDPM \cite{DDPM_2020} with robustness capabilities and is applicable to any DDPM-based diffusion model.
\subsection{Reconstruction-Based Anomaly Segmentation}
Anomaly segmentation is a fundamental task in computer vision. One of the main approaches to address this problem has been matrix decomposition techniques. \cite{RPCA} proposed Robust Principal Component Analysis (RPCA) which decomposes an anomalous image into a low-rank normal background and sparse anomalies. \cite{SSD} introduced Smooth-Sparse Decomposition (SSD) by imposing smoothness on the normal background. These methods, however, are not able to deal with nonlinear patterns. Recently deep generative models have been widely used for anomaly detection and segmentaion \cite{AE_AD_MRI,AE_AD_Structural_similarity,Efficient_GAN_AD, AD_GAN, AD_GAN_anogan_2017, ad_GAN_f-anogan_2019,DIFFUSION_AD_dynamic_noise, DTE_DIFFUSION_AD, Real_net_DIFFUSION_AD, wyatt_anoddpm_2022_DIFFUSION_AD, mousakhan_anomaly_2023_DIFFUSION_AD, zhang_diffusionad_2023_DIFFUSION_AD, iqbal_unsupervised_2023_DIFFUSION_AD, zhang_unsupervised_2023_diffusion_ad, fucka_transfusion_2024_diffusion_ad, imdiffusion_ad_Diffusion}. First, a generative model is trained on normal data and then for an anomalous data, a corresponding normal image is reconstructed such that the difference between the two would create an anomaly segmentation map. Extensive studies have been done with autoencoder and variational autoencoder models \cite{AE_AD_MRI, AD_AE_inpainting, AD_AE_inpainting2, AE_AD_Structural_similarity, AE_memory_AD, Robust_AE_AD, PAedid_ae_ad_memory}. To improve the reconstruction quality in autoencoder-based models, \cite{AE_memory_AD, PAedid_ae_ad_memory} created a memory bank of the embeddings of anomaly free data, which is used as a guide during inference. \cite{AE_AD_Structural_similarity} employed a structural similarity-based loss function different from L2 reconstruction loss to train autoencoders. This loss function incorporates luminance, contrast, and structure of the images. However, autoencoders are known to have reconstruction quality issues \cite{zhang_diffusionad_2023_DIFFUSION_AD, wyatt_anoddpm_2022_DIFFUSION_AD}.\\ GAN models have been extensively studied in image generation \cite{goodfellow_generative_GAN_2014} and  reconstruction-based anomaly segmentation.  \cite{Efficient_GAN_AD, AD_GAN, AD_GAN_anogan_2017, ad_GAN_f-anogan_2019}. For reconstruction, \cite{AD_GAN_anogan_2017} searches for a member in the latent space that can generate the anomaly-free image closest to the input. To improve this approach, \cite{ad_GAN_f-anogan_2019} trained an additional CNN encoder that maps the image space to the latent space. This CNN encoder is then used to map the input to the latent space and reconstruct the anomaly free image using the trained GAN generator. Although GAN models are very powerful, their training is very unstable and sensitive to the choice of parameters and architecture of the network \cite{karras_style-based_2019_GAN_Problem_1, bond-taylor_deep_2022_GAN_problem_3, gulrajani_improved_2017_GAN_problem_4, brock_large_2019_GAN_problem_5, xiao_tackling_2022_GAN_problem_6, arjovsky_wasserstein_2017_GAN_problem_7}. One of the major problems with GAN is mode collapse \cite{zhao_bias_2018_GAN_problem_2, brock_large_2019_GAN_problem_5}, which occurs when the model gets stuck in some portion of the image space and fails to capture mode diversity. \cite{brock_large_2019_GAN_problem_5} suggested monitoring top three singular values for delaying mode collapse in either the generator or discriminator. \\
In recent years, the vast theoretical work behind diffusion models has led to unprecedented success in image synthesis and mode coverage. They have been widely explored in reconstruction-based anomaly segmentation \cite{DIFFUSION_AD_dynamic_noise, DTE_DIFFUSION_AD, Real_net_DIFFUSION_AD, wyatt_anoddpm_2022_DIFFUSION_AD, mousakhan_anomaly_2023_DIFFUSION_AD, zhang_diffusionad_2023_DIFFUSION_AD, iqbal_unsupervised_2023_DIFFUSION_AD, zhang_unsupervised_2023_diffusion_ad, fucka_transfusion_2024_diffusion_ad, imdiffusion_ad_Diffusion}. \cite{wyatt_anoddpm_2022_DIFFUSION_AD} used simplex noise instead of Gaussian noise in DDPM to improve fidelity. \cite{DIFFUSION_AD_dynamic_noise} used the average distance between the input image features and the $k$-nearest neighbor features in the training set to adjust the level of added noise. During inference, if the data point appears distant from the training data, it is subjected to more noise in the forward diffusion process by sampling a larger time step. This ensures that anomalous pixels are effectively corrupted. Another approach to improve the reconstruction fidelity is to use guidance \cite{DTE_DIFFUSION_AD, zhang_diffusionad_2023_DIFFUSION_AD, zhang_unsupervised_2023_diffusion_ad}. \cite{DTE_DIFFUSION_AD} utilizes the input image as a guide to reconstruct the anomaly-free image closest to the input. \cite{Real_net_DIFFUSION_AD, fucka_transfusion_2024_diffusion_ad} generate high-quality synthetic anomalies to improve performance. \cite{mask_ddpm_2023} employed masking for data augmentation, while \cite{imdiffusion_ad_Diffusion} leverages intermediate steps of backward diffusion for more accurate anomaly detection.\\
Reconstruction-based anomaly segmentation, relies on the assumption of having access to anomaly-free training samples. However, this assumption is not true in many application domains including manufacturing and biomedical. \cite{Robust_AE_AD} used an autoencoder as the low-rank component in Robust Principal Component Analysis (RPCA) \cite{RPCA} to improve robustness. That being said, there has been no concrete work on robust diffusion-based anomaly segmentation.\\
Our work focuses on making diffusion-based reconstruction model robust to contamination. This method can be easily generalized to various diffusion-based models. 
\subsection{Robust Regression}
A regression problem can be formulated as \cref{eq:regression} 
where $\mathcal{L}$ is the loss function, $D=\{x_i,y_i\}_{i=1}^N$ represents $N$ training data points, $F_{\theta}$ is the predictor with parameters $\theta$, and $\hat{\theta}$ denotes the optimal parameters of the predictor. 
\begin{equation}
\label{eq:regression}
\hat{\theta} = \arg\min_{\theta} \sum_{i=1}^{N} \mathcal{L}(y_i - F_{\theta}(x_i))
\end{equation}
In this setting, $e_i=r_i=y_i-F_{\theta}(x_i)$ denotes the residual or error for data point $i$. When the predictor errors follow a Laplcian distribution, the Maximum Likelihood Estimation (MLE) leads to minimizing the L1 loss \cite{meyer_alternative_2021}. The Laplacian distribution is better suited for modeling anomalies due to its heavier tails which provide higher probabilities compared to the normal distribution. As shown in \cite{lasso_robust_2010}, L1 norm regularization in Lasso regression \cite{tibshirani1996regression_LASSO} exhibits robustness properties. 
Additionally, Least Trimmed Squares (LTS) \cite{rousseeuw_robust_1987} also has robustness properties. LTS is trained by removing the samples with large residuals using \cref{eq:LTS}. In this formula, $s<N$ is a hyper-parameter that specifies the number of training samples used, and $r_i(\theta) \forall i \in {1,...,N}$ represents the residuals in ascending order.
\begin{equation}
\label{eq:LTS}
\hat{\theta}_{LTS} := \arg\min_{\theta} \sum_{i=1}^{s} r_{[i]}^2(\theta)
\end{equation}
 As noted in \cite{huber_robust_1964} if the error follows a Huber distribution, minimizing the Huber loss is equivalent to Maximum Likelihood Estimation (MLE) . The Huber distribution is a combination of both Laplacian and Gaussian distributions. Also \cite{meyer_alternative_2021} demonstrated that even if the error does not follow the Huber density, Huber loss minimizes the Kullback–Leibler (KL) divergence between model and predictor uncertainty in the presence of contamination. The Huber loss with $\delta$ as the hyper-parameter and $r$ as the residual, is defined in \cref{eq: huber}. 
 \begin{equation}
 \label{eq: huber}
     \text{where} \quad \text{Huber}_\delta(r) = 
\begin{cases} 
    \frac{1}{2}r^2 & \text{if } |r| \leq \delta \\
    \delta \left( |r| - \frac{1}{2}\delta \right) & \text{if } |r| > \delta
\end{cases}
 \end{equation}
In this work, we employ both the Huber loss and the least trimmed squares method to develop our Robust Denoising Diffusion Probabilistic Model (RDDPM).

\section{Methodology}
\label{sec:method} 
\subsection{Background}
\begin{figure}[t]
    \centering
    \includegraphics[width=\columnwidth]{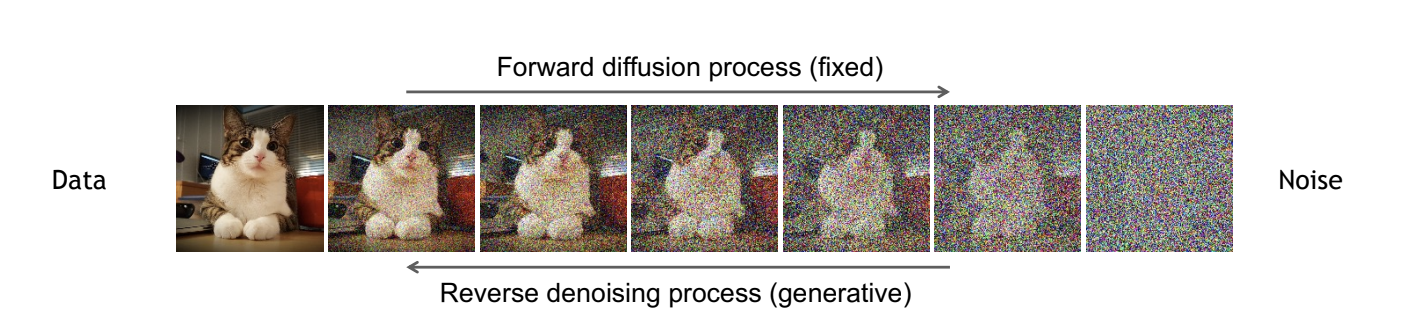}
    \caption{Forward and backward diffusion.}
    \label{fig:DDPM_vis}
\end{figure}

In Denoising Diffusion Probabilistic Models (DDPM) (\cite{DDPM_2020}), a Markov chain is designed to add noise incrementally to the input data and transform the original distribution to the noise distribution e.g., a standard Gaussian distribution. The forward and backward diffusion processes are visualized in \cref{fig:DDPM_vis}, adapted from \cite{cvpr2023diffusion}. In the forward diffusion process, Gaussian noise is added to the data  $\mathbf{x}_0$ over $T$ time steps. In practice, number of time steps is usually chosen to be 1000. Let $\beta_t$ be the noise schedule, where $\beta_t \in (0, 1)$. This parameter is usually chosen to vary linearly between 0.001 and 0.02. The forward process is defined in \cref{eq:forward_ddpm} such that $t \in {1,...,T}$.

\begin{figure}[t]
    \centering
    \includegraphics[width=0.48\textwidth]{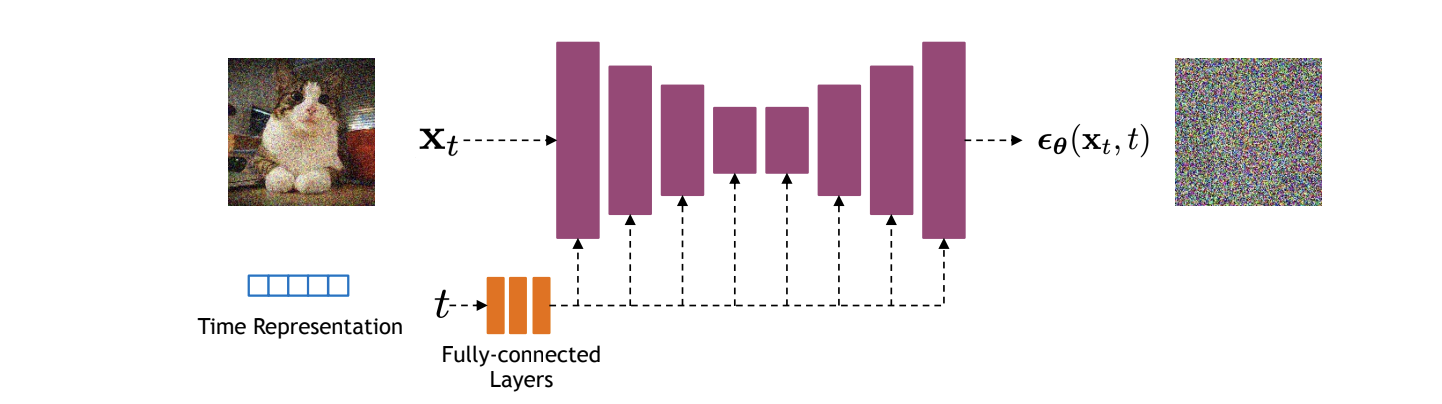}
    \caption{Diffusion model as noise prediction model.}
    \label{fig:noise_pred}
\end{figure}

\begin{equation}
\label{eq:forward_ddpm}
q(\mathbf{x}_t | \mathbf{x}_{t-1}) = \mathcal{N}(\mathbf{x}_t; \sqrt{1 - \beta_t} \mathbf{x}_{t-1}, \beta_t \mathbf{I}),
\end{equation}
By defining \(\alpha_t = 1 - \beta_t\) and the cumulative product \(\bar{\alpha}_t = \prod_{i=1}^t \alpha_i\), the conditional probability distribution at any time step conditioned on time step 0, has an analytical form as given in \cref{eq:any_t}.
\begin{equation}
\label{eq:any_t}
\begin{aligned}
\centering
&q(\mathbf{x}_t | \mathbf{x}_0) = \mathcal{N}(\mathbf{x}_t; \sqrt{\bar{\alpha}_t} \mathbf{x}_0, (1 - \bar{\alpha}_t) \mathbf{I})\\
&\Rightarrow x_t=\sqrt{\bar{\alpha}_t} \mathbf{x}_0+\sqrt{(1 - \bar{\alpha}_t)}z,  z \sim \mathcal{N}(0,\mathbf{I})
\end{aligned} 
\end{equation}
When training, the goal is to learn the reverse conditional distributions $p_\theta(\mathbf{x}_{t-1} | \mathbf{x}_t)=\mathcal{N}(\mathbf{x}_t; \mu(x_t,t), \Sigma(x_t,t))$. The training is performed by maximizing the log likelihood of the data, while considering the reverse process variance constant and modeling the mean as a function of added Gaussian noise \cite{DDPM_2020} in \cref{eq:any_t}. Maximizing log likelihood turns to \cref{eq: Lsimple}.
\begin{equation}
\label{eq: Lsimple}
L_{\text{simple}}(\theta) := \mathbb{E}_{t, \mathbf{x}_0, \boldsymbol{\epsilon}} \left[ 
    \left\| \boldsymbol{\epsilon} - \epsilon_\theta \left( 
        \sqrt{\bar{\alpha}_t} \mathbf{x}_0 + \sqrt{1 - \bar{\alpha}_t} \boldsymbol{\epsilon}, t 
    \right) \right\|^2 
\right]
\end{equation}
in which $\theta, t, x_0, \epsilon$ are model parameters, time step in the forward diffusion, a data point, and a standard Gaussian noise, respectively. \cref{fig:noise_pred}, adapted from \cite{cvpr2023diffusion}, illustrates how diffusion models function as noise prediction models, typically employing a U-Net architecture.\\

In our approach, DDPM is employed for anomaly detection, following the methodology outlined by \cite{anoddpm_2022}. A function  $\epsilon_\theta(x)$ is trained to take a noisy image  $x$  as input and predict the added noise using the loss function defined in \cref{eq: Lsimple}. During inference, for a new anomalous image, noise is added for 25\% of the training steps (e.g., 250 steps) in the forward diffusion process. This ensures that the anomalies are corrupted, and the image becomes close to Gaussian noise, but not too close, so the reconstructed image would resemble the anomalous image. Then, the previous time steps are iteratively sampled using the trained  $\epsilon_{\theta}(x)$  function until reaching step 1, resulting in the reconstruction of the anomaly-free image. The anomaly heatmap is then generated by computing the absolute difference between the reconstructed normal image and the input anomalous image.
During each iteration of training DDPM, a time step, a normal Gaussian noise, and a data point are sampled. The noise is then added to the image according to the predefined noise schedule. Using the noisy image and the corresponding ground truth noise, the parameters of the noise prediction model are updated through gradient descent optimization algorithm. The details of the training and sampling algorithms can be found in
algorithm \cref{alg:train_DDPM} and \cref{alg:sample_DDPM} from \cite{DDPM_2020}.
\begin{algorithm}
\caption{Training algorithm for DDPM}\label{alg:train_DDPM}
\begin{algorithmic}[1]
 \WHILE{Not converged}
    \STATE $x_0 \sim q(x_0)$
    \STATE $t \sim \text{Uniform}(\{1,\dots,T\})$
    \STATE $\epsilon \sim \mathcal{N}(0, I)$
    \STATE Take gradient descent step on 
    \[
    \nabla_\theta \left\| \epsilon - \epsilon_\theta \left(\sqrt{\bar{\alpha}_t}x_0 + \sqrt{1 - \bar{\alpha}_t}\epsilon, t\right) \right\|^2
    \]
\ENDWHILE
\end{algorithmic}
\end{algorithm}
\begin{algorithm}
\caption{Sampling algorithm for DDPM}\label{alg:sample_DDPM}
\begin{algorithmic}[1]
\STATE Sample $x_T \sim \mathcal N(0, I)$
\FOR{$t = T,\dots,1$}
    \STATE Sample $z \sim \mathcal{N}(0, I)$ if $t > 1$, else $z = 0$
    \STATE $x_{t-1} = \frac{1}{\sqrt{\alpha_t}} \left( x_t - \frac{1 - \alpha_t}{\sqrt{1 - \bar{\alpha}_t}} \epsilon_\theta(x_t, t) \right) + \sigma_t z$
\ENDFOR
\STATE Return $x_0$ 
\end{algorithmic}
\end{algorithm}
\subsection{Generative Modeling as Supervised Nonlinear Regression}
Suppose the nonlinear regression in \cref{eq:nonlinear} where $e$ is the prediction error, $\epsilon$ is the sampled Gaussian noise, and $F_{\theta}$ is the noise prediction model in \cref{eq: Lsimple}. We consider the response variable as the sampled Gaussian noise $\epsilon$ , the independent variable as the vector of noisy image in forward diffusion and the time step, and the predictor function $F$ as the noise prediction neural network $\epsilon_\theta$. 
\begin{equation}
\label{eq:nonlinear}
\begin{aligned}
&y=F_{\theta}(x)+e, x=\left( 
        \sqrt{\bar{\alpha}_t} \mathbf{x}_0 + \sqrt{1 - \bar{\alpha}_t} \boldsymbol{\epsilon}, t 
    \right)
\end{aligned}
\end{equation}
Suppose there are $N$ samples. Each sample consists of $(\epsilon_i \sim N(0, I), t_i \sim Uniform [1,T], x_{i0} \sim q(x))$.
As shown in \cite{meyer_alternative_2021}, when the error is drawn from a zero mean Gaussian distribution, the Maximum Likelihood Estimation (MLE) is equivalent to minimizing the L2 loss. As demonstrated in \cref{eq:MLE}, by applying the weak law of large numbers, the L2 loss transforms to the DDPM training loss function in \cref{eq: Lsimple}. This connection allows us to leverage the extensive literature on robust regression techniques for DDPM training. 
\begin{equation}
\label{eq:MLE}
\begin{aligned}
    & \hat{\theta}_{MLE}:= arg min_\theta \Sigma_{i=1}^{N} (y_i - F_\theta (x_i))^2\\& =arg min_\theta \frac{\Sigma_{i=1}^{N} (y_i - F_\theta (x_i))^2 }{N}\\
    &\text{If} \quad N\rightarrow\infty: \frac{\Sigma_{i=1}^{N} (y_i - F_\theta (x_i))^2 }{N}\\&=E_{t,x_0,\epsilon}[(y_i -F_\theta (x_i))^2] =L_{simple}(\theta)\\
    & \text{such that} \quad y_i=\epsilon_i, x_i=\left( 
        \sqrt{\bar{\alpha}_t} \mathbf{x_i}_0 + \sqrt{1 - \bar{\alpha}_{t_i}} \boldsymbol{\epsilon_i}, t_i 
    \right), F=\epsilon_\theta
    \end{aligned}
\end{equation}
\subsection{RDDPM: Robust DDPM}
As outlined in the experiments in \cref{sec:experimens}, diffusion models are not robust to outliers in the training set. When using L2 norm loss in DDPM training, anomalous data points can have a significant impact on the model parameters due to the quadratic nature of the loss, which amplifies the effect of the outliers. In contrast, with RDDPM-Huber or RDDPM-LTS, large deviations have a linear effect or are excluded from training, reducing their impact on the model.
\subsubsection*{RDDPM-LTS}
In the presence of data contamination, one approach to making DDPM robust to outliers is the Least Trimmed Squares (LTS). As shown in \cite{rousseeuw_robust_1987}, LTS learns the regression parameters by considering only the smallest residuals out of the total N residuals when they are arranged in ascending order. The LTS optimization problem is defined in \cref{eq:LTS}. The robust DDPM can be derived by replacing the gradient descent step in \cref{alg:train_DDPM} with \cref{eq:RDDPM-LTS} where $\lambda=\frac{s}{B}$ is the robustness parameter and $B$ is the batch size. 
\begin{equation}
\label{eq:RDDPM-LTS}
\sum_{i=1}^{s=\lambda \times B} \nabla_{\theta} \left\lVert \epsilon_i - \epsilon_{\theta}\left( \sqrt{\bar{\alpha}_{t_i}} x_{0_i} + \sqrt{1 - \bar{\alpha}_{t_i}} \epsilon_i, t_i \right) \right\rVert^2
\end{equation}
As we increase the robustness parameter, our model loses robustness and become less robust. If we set it equal to 1, this update rule would be the same as the original update rule in \cref{alg:train_DDPM} making our RDDPM-LTS equivalent to DDPM.\\
The training and sampling algorithms for RDDPM-LTS are presented in \cref{alg:DDPM-LTS training} and \cref{alg:sample_DDPM}, respectively.
\subsubsection*{RDDPM-Huber}
Another approach to making DDPM robust is to use Huber loss. Huber loss have been shown to possess robustness properties \cite{huber_robust_1964}. In \cite{meyer_alternative_2021} it was demonstrated that Huber loss minimizes the KL divergence between model uncertainty and predictor uncertainty in the case of data contamination. The Huber loss is defined in \cref{eq: huber} where the robustness parameter $\delta$ controls the level of robustness. As $\delta$ increases, the model becomes less robust. Setting $\delta$ to zero turns the Huber loss into L1 loss, while setting it to infinity turns it into L2 loss, making RDDPM-Huber equivalent to DDPM. \\
The training and sampling algorithms for RDDPM-Huber are presented in \cref{alg:RDDPM-Huber training} and \cref{alg:sample_DDPM}, respectively.
\begin{algorithm}
\caption{RDDPM-Huber Training Algorithm}\label{alg:RDDPM-Huber training}
\begin{algorithmic}[1]
\WHILE{Not converged}
    \STATE $x_0 \sim q(x_0)$
    \STATE $t \sim \text{Uniform}(\{1,\dots,T\})$
    \STATE $\epsilon \sim \mathcal{N}(0, I)$
\STATE Take gradient descent step on 
\[
\nabla_\theta \, \text{Huber}_\delta \left(\epsilon - \epsilon_\theta \left(\sqrt{\bar{\alpha}_t}x_0 + \sqrt{1 - \bar{\alpha}_t}\epsilon, t\right) \right)
\]\[
\text{where} \quad \text{Huber}_\delta(r) = 
\begin{cases} 
    \frac{1}{2}r^2 & \text{if } |r| \leq \delta \\
    \delta \left( |r| - \frac{1}{2}\delta \right) & \text{if } |r| > \delta
\end{cases}
\]
\ENDWHILE
\end{algorithmic}
\end{algorithm}
\begin{algorithm}
\caption{RDDPM-LTS Training Algorithm}\label{alg:DDPM-LTS training}
\begin{algorithmic}[1]
 \WHILE{Not converged}
    \STATE $x_0 \sim q(x_0)$
    \STATE $t \sim \text{Uniform}(\{1,\dots,T\})$
    \STATE $\epsilon \sim \mathcal{N}(0, I)$
    \STATE Take gradient descent step on 
    \[
    \nabla_\theta LTS(\left\| \epsilon - \epsilon_\theta \left(\sqrt{\bar{\alpha}_t}x_0 + \sqrt{1 - \bar{\alpha}_t}\epsilon, t\right) \right\|^2
    )\] \[
    = \sum_{i=1}^{s=\lambda \times B} \nabla_{\theta} \left\lVert \epsilon_i - \epsilon_{\theta}\left( \sqrt{\bar{\alpha}_{t_i}} x_{0_i} + \sqrt{1 - \bar{\alpha}_{t_i}} \epsilon_i, t_i \right) \right\rVert^2\]
    \[\text{Where} \ s \in \{1,...,B\} \quad \text{and} \quad \lambda \in (0,1]\]
\ENDWHILE
\end{algorithmic}
\end{algorithm}
In summary, both RDDPM-Huber and RDDPM-LTS can be viewed as generalizations of DDPM, where setting the robustness parameters to 1 for RDDPM-LTS or infinity for RDDPM-Huber recovers the DDPM algorithm.
\section{Experiments}
\label{sec:experimens}
\subsection{Experimental Setup}
\subsubsection*{Datasets}
We validate RDDPM on the challenging high-resolution MVTec Anomaly Detection dataset~\cite{bergmann_mvtec_2019}, which consists of 5,354 RGB images at a resolution of $1024 \times 1024$. The dataset includes 4,096 defect-free images and 1,258 defective images. The anomalies span 73 different defect types across 5 texture categories and 10 object categories, totaling 15 categories in all.

For training, we use the 4,096 defect-free images along with 957 defective images. For the in-domain test set, we evaluate on 209 defective images that contain anomaly types seen during training. To assess generalization to out-of-distribution (OOD) anomalies, we additionally reserve 92 defective images from 5 defect types across 5 categories that are not present in the training data.

We also conduct two focused case studies using specific classes from the MVTec dataset: \textit{carpet} and \textit{grid}. These categories exhibit complex textures and a diverse range of anomalies. The \textit{carpet} class contains 280 normal training images and 89 defective images spanning 5 defect types. The \textit{grid} class includes 264 normal training images and 57 defective images, also across 5 defect types.
\subsubsection*{Implementation Details}
For the noise prediction network, we adopt the architecture described in~\cite{stable_diff_high-resolution_2022}. All training images are resized to $100 \times 100$ resolution. In our experiment using the full MVTec dataset, the model is trained for 20 epochs directly on these resized images with a batch size of 4. For class-specific experiments, we divide the training images into $28 \times 28$ patches and use a total of 50{,}000 patches for training for 10 epochs.

To simulate anomaly effects in the data, we apply synthetic corruptions to the patches. Based on a predefined corruption ratio, we randomly select 70\% of the $2 \times 2$ blocks within each $28 \times 28$ patch and multiply their intensities by a factor of 5, emulating measurement artifacts or anomaly-like distortions. Crucially, no defective images are used during training; they are reserved exclusively for evaluation to ensure the models are not exposed to anomalous data during training.

We evaluate our method under corruption levels of 0, 10, 20, and 30. For our RDDPM model, we use Huber loss with a fixed $\delta = 0.2$. 
\subsubsection*{Evaluation Metrics}
Our method, along with all benchmark approaches, is based on reconstruction-based anomaly segmentation, producing a heatmap for each anomalous image. By applying post-processing techniques such as thresholding or domain-specific methods, a binary anomaly mask can be derived from the heatmap. To ensure a fair and general comparison that is independent of specific post-processing choices, we evaluate all methods using pixel-level Area Under the Receiver Operating Characteristic Curve (AUROC) and Area Under the Precision-Recall Curve (AUPRC). Additionally, we report the reconstruction Mean Squared Error (MSE) over non-defective regions to assess the model's ability to accurately reconstruct normal content.
\subsection{Anomaly Segmentation Comparison}
We compare our method against DDPM \cite{DDPM_2020} and two state-of-the-art diffusion-based models for industrial anomaly detection. AnoDDPM \cite{anoddpm_2022} adds noise to the input for 250 diffusion steps and then denoises it using Simplex noise guidance. DiffusionAD\cite{zhang_diffusionad_2023_DIFFUSION_AD} generates two noisy versions of the image at different noise scales during the forward diffusion process. It then reconstructs the image in a single step using the higher noise scale and refines the result conditioned on this prediction using the lower scale. This design ensures improved reconstruction quality through conditional refinement. 

One advantage of our model is that, unlike the other methods, which contribute to image reconstruction only during sampling and backward diffusion, our method integrates into the diffusion training process itself. 

As shown in \cref{tab: AUROC Comparison on MVTec AD dataset}, under 20\% corruption, our method outperforms DDPM on both in-domain and out-of-domain anomalies in the MVTec dataset. Corresponding qualitative results are shown in \cref{fig:DDPM contaminated}.

\begin{table}[htbp]
\centering
\caption{AUROC Comparison on MVTec AD dataset}
\renewcommand{\arraystretch}{1} 
\begin{adjustbox}{max width=\linewidth}
\begin{tabular}{|c|c|c|}

\hline
\textbf{Anomaly kind} & \textbf{DDPM} & \textbf{RDDPM} \\
\hline
In domain anomalies & 0.76 & \textbf{0.78}    \\
\hline
Out of domain anomalies &0.69  & \textbf{0.71}  \\
\hline
\end{tabular}
\end{adjustbox}
\label{tab: AUROC Comparison on MVTec AD dataset}
\end{table}

We also report results across AUROC, AUPRC, and MSE for both the \textit{carpet} and \textit{grid} categories using three methods: RDDPM, AnoDDPM, and DiffusionAD. As shown in \cref{tab:20percent_contamination}, our method consistently outperforms both AnoDDPM and DiffusionAD in terms of AUROC and AUPRC. For example, in the \textit{grid} category, RDDPM achieves 8.08\% higher AUROC and 10.37\% higher AUPRC compared to the second-best method, DiffusionAD. In terms of MSE, RDDPM exhibits a slightly higher reconstruction error than the best-performing method, though the difference remains marginal.

\begin{table}[t]
\centering
\caption{20\% Contamination Results on Carpet and Grid Categories. $\uparrow$ indicates higher is better, $\downarrow$ indicates lower is better.}
\begin{tabular}{lccc}
\toprule
\textbf{Method} & \textbf{AUROC $\uparrow$} & \textbf{AUPRC $\uparrow$} & \textbf{MSE $\downarrow$} \\
\midrule
\multicolumn{4}{c}{\textbf{Carpet}} \\
\midrule
RDDPM       & \textbf{0.5673} & \textbf{0.0362} & 0.1246 \\
AnoDDPM     & 0.4650          & 0.0234          & 0.2115 \\
DiffusionAD & 0.4909          & 0.0268          & \textbf{0.1199} \\
\midrule
\multicolumn{4}{c}{\textbf{Grid}} \\
\midrule
RDDPM       & \textbf{0.6373} & \textbf{0.1803} & 0.0896 \\
AnoDDPM     & 0.4734          & 0.0121          & 0.2188 \\
DiffusionAD & 0.5565          & 0.0766          & \textbf{0.0863} \\
\bottomrule
\end{tabular}
\label{tab:20percent_contamination}
\end{table}
\subsection{Ablation Studies}

\subsubsection*{Robustness Parameter}
We also conduct an experiment to investigate the effect of the robustness parameter on the performance of RDDPM. As discussed earlier, this parameter governs the trade-off between robustness and learning in the Huber loss. When the robustness parameter \(\delta = 0\), the Huber loss reduces to the pure \(\ell_1\) norm. Increasing \(\delta\), on the other hand, makes the loss behave more like the mean squared error (MSE) loss.

It is important to note that the input images are normalized to the range \([-1, 1]\), meaning the values passed to the loss function lie within \([0, 2]\). For example, setting \(\delta = 0.2\) effectively penalizes deviations greater than 10\% of the full intensity range.

\begin{figure}[t]
    \centering
    \begin{minipage}[b]{0.48\textwidth}
        \centering
        \includegraphics[width=\textwidth]{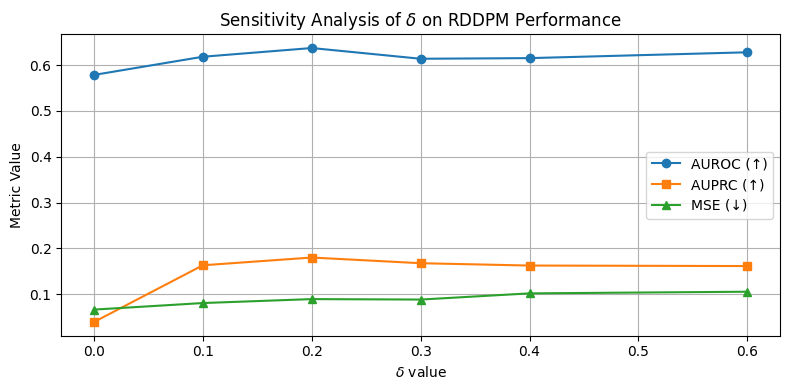} 
        \caption*{(a) Sensitivity of RDDPM performance w.r.t. $\delta$}
    \end{minipage}
    \hfill
    \begin{minipage}[b]{0.48\textwidth}
        \centering
        \captionof{table}{Metric values across different $\delta$ values. $\uparrow$ indicates higher is better, $\downarrow$ lower is better.}
        \label{tab:delta_sensitivity}
        \begin{tabular}{lccccc}
            \toprule
            Metric & 0 & 0.1 & 0.2 & 0.3 & 0.4 \\
            \midrule
            AUROC $\uparrow$ & 0.5786 & 0.6183 & \textbf{0.6373} & 0.6139 & 0.6153 \\
            AUPRC $\uparrow$ & 0.0396 & 0.1634 & \textbf{0.1803} & 0.1679 & 0.1628  \\
            MSE $\downarrow$ & \textbf{0.0667} & 0.0810 & 0.0896 & 0.0886 & 0.1021  \\
            \bottomrule
        \end{tabular}
    \end{minipage}
    \caption{Sensitivity analysis of RDDPM to Huber loss hyperparameter $\delta$.}
    \label{fig:delta_sensitivity}
\end{figure}

We evaluate the performance of RDDPM at several levels of \(\delta\): 0 (0\%), 0.1 (5\%), 0.2 (10\%), 0.3 (15\%), and 0.4 (20\%). The quantitative results are summarized in \cref{tab:delta_sensitivity} and visualized in \cref{fig:delta_sensitivity}. AUROC and AUPRC scores are lowest at \(\delta = 0\), which is expected since the loss function is pure \(\ell_1\) norm. These metrics rise significantly at \(\delta = 0.1\) and \(\delta = 0.2\), reaching their peak at \(\delta = 0.2\), before slightly declining for higher values of \(\delta\), likely due to the reduced robustness.

Interestingly, the lowest MSE occurs when \(\delta = 0\), possibly because the pure \(\ell_1\) loss encourages the model to learn a mean representation across all training images rather than reconstructing each one individually. This averaging effect can result in deceptively low reconstruction error. Further analysis is needed to better understand this phenomenon.

Overall, it can be observed that setting the robustness parameter to any value greater than zero yields consistently high AUROC and AUPRC scores and low MSE, indicating that the performance is largely insensitive to the exact choice of \(\delta\) beyond zero.

\subsubsection*{Corruption Ratio}
\begin{figure*}[t]
    \centering
    \includegraphics[width=\textwidth]{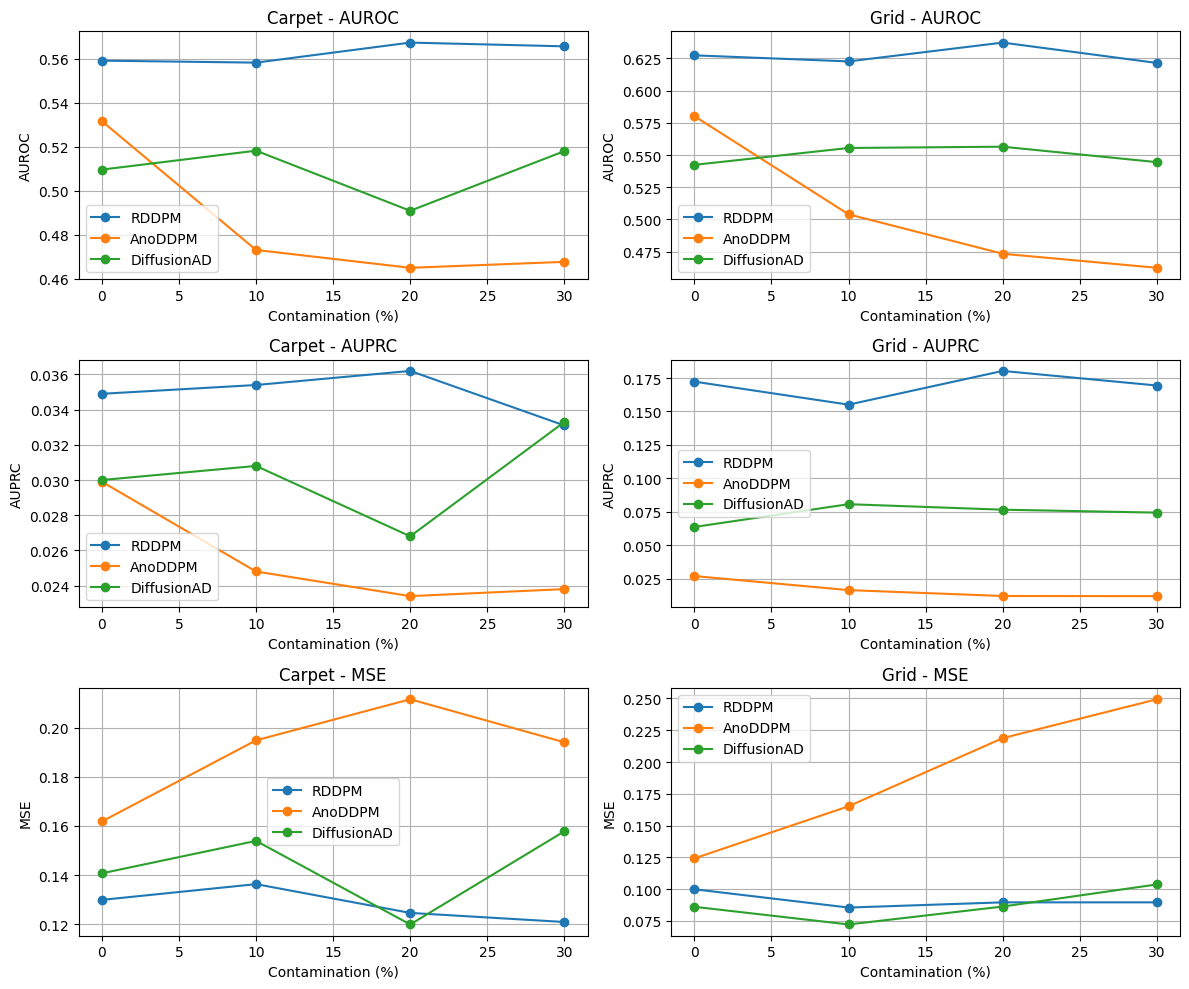}
    \caption{Performance metrics for RDDPM, AnoDDPM, and DiffusionAD across contamination levels.}
    \label{fig:corruption}
\end{figure*}

We also investigate the impact of training data corruption on the performance of all competing methods. As shown in \cref{fig:corruption}, RDDPM consistently outperforms both AnoDDPM and DiffusionAD across all contamination levels ranging from 0\% to 30\% in terms of AUROC and AUPRC. The only exception occurs in the carpet category, where DiffusionAD achieves almost the same AUPRC at 30\% contamination. AnoDDPM consistently underperforms relative to the other methods across all metrics except when there is no corruption. In terms of MSE, RDDPM achieves the lowest reconstruction error in the carpet category, with a slight underperformance only at the 20\% contamination level. In the grid category, RDDPM ranks second overall but outperforms all other methods when the contamination level exceeds 20\%.

Overall, our model consistently demonstrates superior performance across varying contamination levels. Notably, even in the absence of contamination, it achieves stronger anomaly detection capability compared to competing methods.

\section{Conclusion}
\label{sec:conclusion}
In this work, we introduced \textbf{RDDPM}, a robust generalization of the DDPM framework designed to handle outliers in the training data. We proposed two variants of our model, each equipped with tunable parameters to control the robustness-learning trade-off. We evaluated RDDPM on the MVTec AD anomaly detection benchmark and compared it against state-of-the-art diffusion-based anomaly segmentation methods. Our model consistently outperformed existing approaches across varying contamination levels in terms of AUROC, AUPRC, and reconstruction MSE. Furthermore, a sensitivity analysis on the robustness parameter demonstrated that RDDPM maintains stable performance across a wide range of values, with the exception of zero, where learning becomes ineffective due to the change in the loss function.
\FloatBarrier
\clearpage 
{
    \small
    \bibliographystyle{ieeenat_fullname}
    \bibliography{main}
}


\end{document}